\documentclass{IEEEtran4PSCC}

\ifCLASSINFOpdf
   \usepackage[pdftex]{graphicx}
\else
   \usepackage[dvips]{graphicx}
\fi

\usepackage[cmex10]{amsmath}
\usepackage{amssymb}
\usepackage{booktabs}
\ifCLASSOPTIONcompsoc
 \usepackage[caption=false,font=normalsize,labelfont=sf,textfont=sf]{subfig}
\else
 \usepackage[caption=false,font=footnotesize]{subfig}
\fi

\usepackage{xcolor}

\makeatletter
\let\old@ps@headings\ps@headings
\let\old@ps@IEEEtitlepagestyle\ps@IEEEtitlepagestyle
\def\psccfooter#1{%
    \def\ps@headings{%
        \old@ps@headings%
        \def\@oddfoot{\strut\hfill#1\hfill\strut}%
        \def\@evenfoot{\strut\hfill#1\hfill\strut}%
    }%
    \def\ps@IEEEtitlepagestyle{%
        \old@ps@IEEEtitlepagestyle%
        \def\@oddfoot{\strut\hfill#1\hfill\strut}%
        \def\@evenfoot{\strut\hfill#1\hfill\strut}%
    }%
    \ps@headings%
}
\makeatother

\psccfooter{%
        \parbox{\textwidth}{\hrulefill \\ \small{23rd Power Systems Computation Conference} \hfill \begin{minipage}{0.2\textwidth}\centering \vspace*{4pt} \includegraphics[scale=0.06]{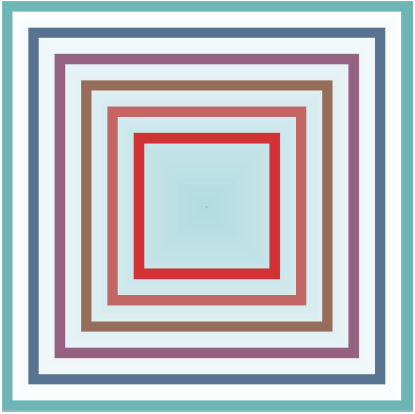}\\\small{PSCC 2024} \end{minipage} \hfill \small{Paris-Saclay, France --- June 4 -- June 7, 2024}}%
}

\begin{document}

\title{Physics-Informed Graph Neural Network for Dynamic Reconfiguration of Power Systems}

\author{\IEEEauthorblockN{Jules Authier\IEEEauthorrefmark{1}\IEEEauthorrefmark{2}
Rabab Haider\IEEEauthorrefmark{1},
Anuradha Annaswamy\IEEEauthorrefmark{1} and 
Florian D\"orfler\IEEEauthorrefmark{2}}
\IEEEauthorblockA{\IEEEauthorrefmark{1}Department of Mechanical Engineering,
MIT, 
Cambridge, USA}
\IEEEauthorblockA{\IEEEauthorrefmark{2}Automatic Control Laboratory,
ETH Zurich,
Zurich, Switzerland}
}

\maketitle

\begin{abstract}
To maintain a reliable grid we need fast decision-making algorithms for complex problems like Dynamic Reconfiguration (DyR). DyR optimizes distribution grid switch settings in real-time to minimize grid losses and dispatches resources to supply loads with available generation. DyR is a mixed-integer problem and can be computationally intractable to solve for large grids and at fast timescales. We propose GraPhyR, a Physics-Informed Graph Neural Network (GNNs) framework tailored for DyR. We incorporate essential operational and connectivity constraints directly within the GNN framework and train it end-to-end. Our results show that GraPhyR is able to learn to optimize the DyR task.
\end{abstract}

\begin{IEEEkeywords}
Graph Neural Network, Dynamic Reconfiguration, Physics Informed Learning.
\end{IEEEkeywords}

\thanksto{\noindent This work is partially supported by the MIT Energy Fellowship.}

\section{Introduction}
The global energy landscape is rapidly evolving with the transition towards renewable energy generation. This transition brings numerous benefits for the climate, but also presents challenges in effectively controlling and optimizing power systems with high penetration of intermittent renewable generation such as solar and wind. New operating schemes are needed to ensure efficient and reliable grid operations in the presence of intermittent generation. Significant research efforts focus on optimizing resource dispatch and load flexibility towards reducing costs and increasing grid efficiency; however there remains efficiency gains to be had when co-optimizing grid topology. To this end, we propose \textit{Dynamic Reconfiguration} (DyR) in a distribution grid to increase operating efficiency by co-optimizing grid topology and resource dispatch.

The distribution grid reconfiguration problem involves the selection of switch states (open/closed) to meet demand with available generation, while satisfying voltage and operating constraints. Grid reconfiguration can re-route power flows to reduce power losses \cite{network_rec}, increase utilization of renewable generation \cite{Lueken_DyR_2012, haider_SiPhyR}, and re-energize grids after contingencies. Presently, DyR is deployed for loss reduction in the EU \cite{Lueken_DyR_2012}, and for fault conditions in the US using rule-based control schemes. The widespread growth of distributed generation, storage, and electric vehicles creates the opportunity for DyR for loss reduction, whereby topology and dispatch decisions are made \textit{fast and frequently} in response to faster resource timescales; as solar generation varies, the topology is adapted to supply loads in close proximity to generation, thus reducing losses and improving voltage profiles across the grid.

The DyR problem is a mixed integer program (MIP) due to the discrete nature of switch decisions. It is well known that MIPs are NP-hard (i.e. cannot be solved in polynomial time) and thus may be computationally intractable for large-scale problems. A distribution substation may have 10 feeders each with 5 switches, resulting in over $10^{15}$ possible topologies. If operating constraints and load conditions result in only $1\%$ of these topologies, the search space remains prohibitively large for traditional approaches. One option is to restrict the optimization to a single feeder, however optimizing topology over all feeders permits load transfer and generation exports. 

Machine learning (ML) offers an alternative by shifting the computational burden to offline training, thereby making dynamic decision making via the online application of ML algorithms computationally feasible. Recent works propose ML for solving MIPs and combinatorial optimization (CO) \cite{Nair2020SolvingNetworks}, either in an end-to-end fashion or to accelerate traditional solvers. Graphs play a central role in formulating many CO problems \cite{Cappart2021CombinatorialNetworks}, representing paths between entities in routing problems, or interactions between variables and constraints in a general CO \cite{zhang2023survey, GuptaHybridBranch}. The use of Graph Neural Networks (GNNs) is also being explored to leverage the underlying graph structure during training and identify common patterns in problem instances. The traveling salesman problem (TSP) is a fundamental problem in CO and a standard benchmark which has been extensively studied with traditional optimization techniques. Recently, GNNs have been used to solve the TSP with good performance and generalizability \cite{Joshi2019AnProblem, Joshi2020LearningGeneralization, Prates2018LearningTSP}. In this work we leverage GNNs to learn the power flow representation for reconfiguration.

Grid reconfiguration for distribution grids has been studied with varying solution methodologies including knowledge-based algorithms and single loop optimization \cite{network_rec,rec_single_loop}, heuristic methods \cite{Wang2022DistributionAlgorithm, 1626402}, and reformulation as a convex optimization problem using big-$\mathcal{M}$ constraints \cite{Taylor2012ConvexReconfiguration,Kovacki2018ScalableApproach, Popovic2022Multi-periodAutomation}. However, these methods are not computationally tractable for large-scale optimization in close to real-time applications, and may be limited to passive grids (i.e. no local generation). Machine learning approaches for DyR have also been proposed \cite{Kundacina2022SolvingLearning,haider_SiPhyR}. In \cite{Kundacina2022SolvingLearning} the DyR problem is formulated as a Markov decision process and solved using reinforcement learning. In \cite{haider_SiPhyR} a light-weight physics-informed neural network is proposed as an end-to-end learning to optimize framework with guarantee certified satisfiability of the power physics. A physics-informed rounding layer explicitly embeds the discrete decisions within the neural framework. These approaches show potential, but both are limited to a given grid topology and switch locations. Our approach is similar to that of \cite{haider_SiPhyR} wherein we embed discrete decisions directly within an ML framework.

The main contribution of this paper is \textbf{GraPhyR}, a graph neural network (\textbf{Gra}) framework employing physics-informed rounding \cite{haider_SiPhyR} (\textbf{PhyR}) for DyR in distribution grids. GraPhyR is an end-to-end framework that learns to optimize the reconfiguration task, enabled by four key architectural components:

\noindent \textbf{(1) A message passing layer that models switches as gates:} The gates are implemented as a value between zero and one to model switches over a continuous operating range. Gates control the flow of information through switches in the GNN, modeling the control of physical power flow between nodes.

\noindent \textbf{(2) A scalable local prediction method:} We make power flow predictions locally at every node in the grid using local features. The predictors are scale-free and so can generalize to grids of any topology and size.

\noindent \textbf{(3) A physics-informed rounding layer:} We embed the discrete open/closed decisions of switches directly within the neural framework. PhyR selects a grid topology for each training instance upon which GraPhyR predicts a feasible power flow and learns to optimize a given objective function.

\noindent \textbf{(4) A GNN that takes the electrical grid topology as input:} We treat the grid topology and switch locations as an input which permits GraPhyR to learn the power flow representation across multiple possible distribution grid topologies within and across grids. Thus GraPhyR can optimize topology and generator dispatch: (a) on multiple grid topologies seen during training, and (b) under varying grid conditions such as (un)planned maintenance of the grid.

We demonstrate the performance of GraPhyR in predicting near-optimal and feasible solutions. We also show the effectiveness of GraPhyR in adapting to unforeseen grid conditions. 

The remainder of this paper is organized as follows. Section~\ref{sec:reconfig_opt} presents DyR as an optimization problem. Section~\ref{sec:graphyr} presents the GraPhyR method and details the four key architectural components. Section~\ref{sec:results} presents the simulation results, and conclusions are drawn in Section~\ref{sec:conclusion}.

\section{Reconfiguration as an Optimization Problem} \label{sec:reconfig_opt}
We consider DyR of distribution grids with high penetration of distributed generation. We model the power physics using Linearized DistFlow \cite{network_rec} as below:
\begingroup
\allowdisplaybreaks
\begin{align}
     \min_{\boldsymbol{\psi}} & \;  f(\mathbf{x}, \boldsymbol{\psi}) = \sum_{\scriptscriptstyle (i,j) \in \mathcal{A}} (p_{ij}^2 + q_{ij}^2 )R_{ij} &\label{eq:objectivefnc}\\
     \text{s.t. } &p_j^G - p_j^L = \sum_{\scriptscriptstyle k: (j,k) \in \mathcal{A}\cup\mathcal{A}{sw}}\hspace{-10pt}p_{jk} - \hspace{-10pt}\sum_{\scriptscriptstyle i:(i,j) \in \mathcal{A}\cup\mathcal{A}{sw}}\hspace{-10pt}p_{ij}, \quad\forall j \in \mathcal{N}\label{eq:pbalance}\\ 
    &q_j^G - q_j^L = \sum_{\scriptscriptstyle k: (j,k) \in \mathcal{A}\cup\mathcal{A}{sw}}\hspace{-10pt}q_{jk} - \hspace{-10pt}\sum_{\scriptscriptstyle i:(i,j)\in\mathcal{A}\cup\mathcal{A}{sw}}\hspace{-10pt}q_{ij}, \quad\;\forall j \in\mathcal{N}\label{eq:qbalance} \\
    &v_i - v_j = 2(R_{ij}p_{ij} + X_{ij}q_{ij}), \qquad\quad\, \forall (i,j)\in\mathcal{A}\label{eq:Ohm}\\
    &\hspace{-20pt}{\begin{cases}
        v_i - v_j = 2(R_{ij}p_{ij} + X_{ij}q_{ij}) \text{ if } y_{ij} = 1\\
        \text{inactive} \hspace{93pt}\text{if } y_{ij} = 0 
    \end{cases} \hspace{-9pt}\forall(i,j)\in \mathcal{A}_{sw}} \label{eq:Ohm_sw}\\
    -&\mathcal{M} y_{ij} \leq p_{ij} \leq \mathcal{M}y_{ij}, \hspace{66pt} \forall (i,j)\in \mathcal{A}_{sw} \label{eq:sw_p}\\
    -&\mathcal{M} y_{ij} \leq q_{ij} \leq \mathcal{M}y_{ij}, \hspace{66pt} \forall (i,j)\in \mathcal{A}_{sw} \label{eq:sw_q}\\
    &p^G_j \in \left[\underline{p}^G_j,\overline{p}^G_j\right],\; q^G_j \in \left[\underline{q}^G_j,\overline{q}^G_j\right] \qquad\qquad\;\,\forall j\in\mathcal{N}\label{eq:var_lims}\\
    &v_j \in \left[\underline{v}, \overline{v}\right],\; v_{j^\#} = 1, \hspace{88pt}  \forall j\in\mathcal{N}\label{eq:v_pcc}\\
    &\sum_{\scriptscriptstyle (i,j)\in \mathcal{A}_{sw}} y_{ij} = N-1-M \label{eq:radial} \\
    &|\delta_{\scriptscriptstyle \mathcal{A}}(j)| + \sum_{\scriptscriptstyle j:(i,j)\in A_{sw}} \hspace{-5pt} y_{ij} + \sum_{\scriptscriptstyle j:(j,i)\in A_{sw}} \hspace{-5pt} y_{ji} \geq 1, \;\;\,\forall j\in\mathcal{N} \label{eq:connectivity}
\end{align}
where $\mathbf{x} = [\mathbf{p^L}, \mathbf{q^L}]$ and the decision variable is ${\boldsymbol{\psi}} = [\mathbf{y}, \mathbf{v}, \mathbf{p_{ij}}, \mathbf{q_{ij}}, \mathbf{p^G}, \mathbf{q^G}]$. 
The equations above describe a distribution grid as a directed graph $\mathcal{G}(\mathcal{N}, \mathcal{A}, \mathcal{A}_{sw})$, with $\mathcal{N}$ the set of $N$ nodes, $\mathcal{A}$ the set of $M$ directed edges, and $\mathcal{A}_{sw}$ the set of $M_{sw}$ switches. The distribution grid is connected to the transmission grid via the point of common coupling (PCC), node $j^\#$, which is the slack bus. The real and reactive power loads at a node $i$ are $p_i^L, q_i^L$, and generation are $p_i^G, q_i^G$. Generation at the PCC indicates import from the transmission grid. The squared magnitude of the voltage at node $i$ is $v_i$. The directed real and reactive power flows across a line (switch) from node $i$ to $j$ are $p_{ij}$ and $q_{ij}$. The power flows are uniquely defined on the directed graph, where $p_{ij} > 0$ indicates flow from $i$ to $j$, and $p_{ij} < 0$ indicates flow from $j$ to $i$. The same applies for $q_{ij}$. The switch status is given by $y_{ij}, \forall (i,j) \in A_{sw}$ and takes on a binary value, one if the switch is closed and zero is the switch is open. The line resistance and reactance are $R_{ij}$ and $X_{ij}$ respectively. We define $\delta_\mathcal{A}(j)$ as the set of edges in $\mathcal{A}$ connected to node $j$, and $|\delta_{\scriptscriptstyle \mathcal{A}}(j)|$ as the number of those edges.

The objective function in \eqref{eq:objectivefnc} linearly approximates electric losses. Eq.~\eqref{eq:pbalance}-\eqref{eq:Ohm_sw} describe the Linearized DistFlow model \cite{network_rec} which assumes lossless power balance \eqref{eq:pbalance}-\eqref{eq:qbalance}, and approximates Ohm's Law as a linear relationship between voltages and power \eqref{eq:Ohm}-\eqref{eq:Ohm_sw}. Eq.~\eqref{eq:Ohm_sw} accommodates switches in the model with a conditional constraint where Ohm's Law is enforced for closed switches. The big-$\mathcal{M}$ constraint in ~\eqref{eq:sw_p} and \eqref{eq:sw_q} enforces power flows through open switches to be zero. Eq.~\eqref{eq:var_lims} describes the nodal injection constraints. Eq.~\eqref{eq:v_pcc} sets the voltage constraints and the slack bus voltage. Eq.~\eqref{eq:radial}-\eqref{eq:connectivity} describe radiality and connectivity constraints required for distribution grids under normal operations. We assume existing protection schemes are used which require radiality of the grid topology; we then enforce radiality in the reconfiguration problem. Note that \eqref{eq:connectivity} is not sufficient to enforce connectivity. To maintain a simple problem formulation, we leverage the fact that the power flow constraints implicitly enforce connectivity: a load cannot be supplied if it is disconnected from the grid.

\section{GraPhyR: End-to-end learning for dynamic reconfiguration}\label{sec:graphyr}
We propose GraPhyR, a physics-informed machine learning framework to solve \eqref{eq:objectivefnc}-\eqref{eq:connectivity}. Our framework in Fig.~\ref{fig_graphyr} features four architectural components: (A) gated message passing to model switches, (B) local predictions to scale across nodes, (C) physics-informed rounding to handle binary variables, and (A) topology input data for adaptability during online deployment. We embed the physics of the distribution grid and reconfiguration problem within each component of the GraPhyR framework. First, the GNN embeds the topology of the underlying distribution grid, and explicitly models the switches using gated message passing. Second, the topology selection embeds the discrete open/close decision of the switches using the physics-informed rounding. Third, we use the power flow equations to predict a subset of variables (denoted as the independent variables), and compute the remaining variables in a recovery step. The GraPhyR framework uses these physics-informed layers to learn to optimize the reconfiguration task while satisfying equality and binarity constraints. The framework is presented in detail next.

\begin{figure*}[!ht]
\centering
\includegraphics[width=5.5in,trim={0 0 0 0}, clip]{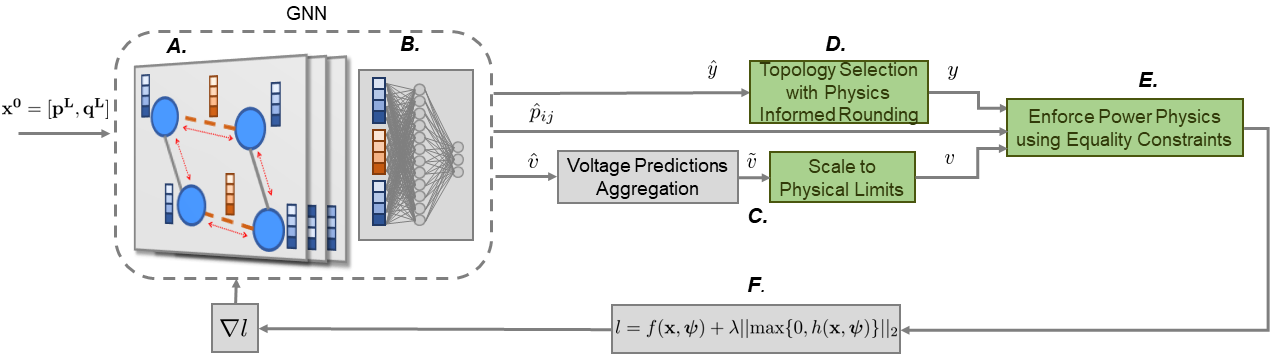}
\caption{GraPhyR: proposed framework to solve the DyR problem.}
\label{fig_graphyr}
\end{figure*}

\subsection{Message Passing}
The GNN models the distribution grid topology as an undirected graph, with switch embeddings modeling the switches in the electrical grid. The GNN's message passing layers incorporate these embeddings as gates, which enables GraPhyR to learn the representation of linearized Ohm's law of \eqref{eq:Ohm_sw} across multiple topologies in a physics-informed way. The input to the GNN are the grid topology and nodal loads, and the output is a set of node and switch embeddings which will be used to make reconfiguration, power flow, and voltage predictions.

\subsubsection{Grid Topology as Input Data for Graph Structure}
An input to the GNN is the grid topology described by $\mathcal{G}(\mathcal{N}, \mathcal{A}, \mathcal{A}_{sw})$, using which the GNN models the physical grid topology as an undirected graph $\mathcal{G}(\mathcal{N}, \mathcal{E},\mathcal{E}_{sw})$ with $N$ nodes, $M$ lines, and $M_{sw}$ switches. Trivially, $\mathcal{E}$ ($\mathcal{E}_{sw}$) represents the undirected communication links along the directed edges $\mathcal{A}$ ($\mathcal{A}_{sw}$) to support message passing and extracting the problem representation in the embeddings. By including $\mathcal{G}(\mathcal{N}, \mathcal{A}, \mathcal{A}_{sw})$ as an input to the GNN, our GraPhyR framework is able to adapt to changing grid conditions, rather than requiring a large training dataset with multiple scenarios.

\subsubsection{Initial Node, Line, and Switch Embeddings}
The second input data to the GNN is the load data $\mathbf{x}^0$ which defines the node embeddings. The load data contains the active and reactive power load  $p^L_i$ and $q^L_i$ for each node $i$ in the grid and thus determines the initial node embeddings $x^0_i$ of every node $i$ in the corresponding graph where $\mathbf{x}^0 = \begin{bmatrix} x_1^0, \hdots, x_N^0 \end{bmatrix}^T = \begin{bmatrix} (p^L_0, q^L_0), \hdots, (p^L_N, q^L_N) \end{bmatrix}^T$. The line embeddings are set to one and are not updated by the message passing layers. The switch embeddings determine the value of the gate and are randomly initialized, similar to randomly initializing weights in a neural network. The switch embeddings are updated through the message passing layers. Initial line and switch embeddings are given by $z_{ij}^0,\; \forall \{i,j\} \in \mathcal{E}\cup\mathcal{E}_{sw}$.

\subsubsection{Message Passing Layers}
In each hidden layer of the GNN the nodes in the graph iteratively aggregate information from their local neighbors. Deeper GNNs have more hidden layers and thus have node embeddings which contain information from further reaches of the graph. For each node embedding $x_i^0$ in the graph, the first message passing layer is defined in \eqref{eq:node0} where $\mathcal{N}_i$ denotes the set of neighboring nodes to node $i$. For each switch embedding $z_{ij}^0$ in the graph, the first message passing layer is defined in \eqref{eq:z0}. 
\begingroup
\begin{align}
    &x_i^{1} = ReLU(W_1^0 x_i^{0} + \sum_{j\in\mathcal{N}_i} \{W_2^0\cdot f(z_{ij}^0) \cdot x_j^{0}\})\label{eq:node0}\\ 
    &f(z_{ij}^0) = \begin{cases}
        sig(z_{ij}^0) \quad \text{if } \{i,j\} \in \mathcal{E}_{sw}\\
        1 \ \quad\qquad \text{otherwise} 
    \end{cases} \label{eq:f0} \\
    &z_{ij}^{1} = 
    ReLU(W_3^0 (x_i^0 + x_j^0) + W_4^0 z_{ij}^0), \forall \{i,j\} \in \mathcal{E}_{sw}\label{eq:z0}
\end{align}
\endgroup
For the remaining message passing layers, denoted by $l \in \{1,2,\dots, \mathcal{L}-1\}$, a residual connection is added to improve prediction performance and training efficiency \cite{residual}. The resulting node and switch embeddings are:
\begingroup
\begin{align}
    &x_i^{l+1} = x_i^l + ReLU(W_1^l x_i^{l} + \sum_{j\in\mathcal{N}_i} \{W_2^l\cdot f(z_{ij}) \cdot x_j^{l}\})\label{eq:node}\\
    &f(z_{ij}^l) = \begin{cases}
        sig(z_{ij}^l) \quad \text{if } \{i,j\} \in \mathcal{E}_{sw}\\
        1 \ \quad\qquad \text{otherwise} \label{eq:f}
    \end{cases}\\
    & z_{ij}^{l+1} = z_{ij}^l + 
    ReLU(W_3^l (x_i^l + x_j^l) + W_4^l z_{ij}^l), \forall \{i,j\} \in \mathcal{E}_{sw}
\end{align}
\endgroup

The line embeddings are trivially set to one.  We omit residual connections in the first message passing layer to expand the input embeddings $x_i^0$ with dimensions of the input data, to an arbitrarily large hidden embeddings dimension $h$. This allows the GNN to learn more complex representations by extracting features in a higher dimensional space.

\subsubsection{Gates} We implement gates in the message passing layer by applying a sigmoid to the switch embeddings, as in \eqref{eq:f0} and \eqref{eq:f}. The function $f(z_{ij})$ acts like a filter for the message passing between two neighboring nodes, attenuating the information signal if the switch is closed. The gate models the switches as a continuous switch (ex. a household light dimmer), controlling information flow in the same way a switch controls power flow between two nodes. 

\subsubsection{Global Graph Information}
In the final message-passing layer, we calculate a global graph embedding, $x_G^\mathcal{L}=\sum_{i=1}^N x_i^\mathcal{L}$. This embedding offers information access across the graph and can reduce the need for an excessive number of message passing layers for sparse graphs, such as those in power systems. This improves the computational efficiency.

\begin{figure}[!ht]
    \centering
    \includegraphics[width=2.7in]{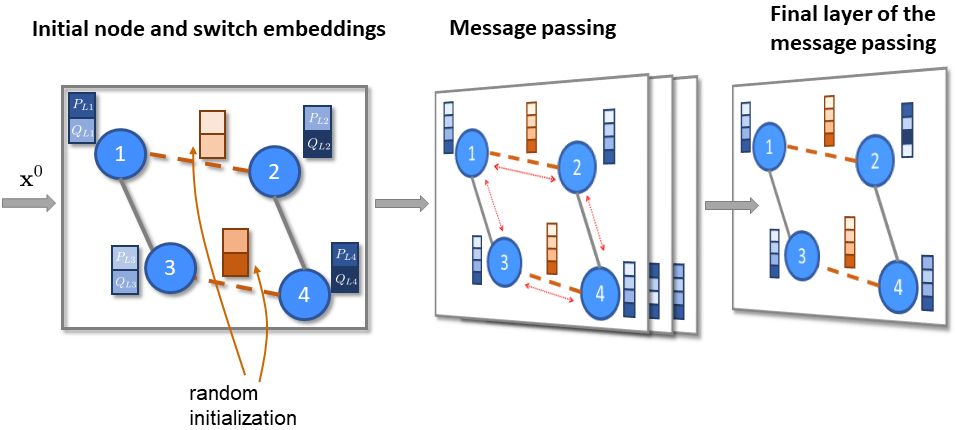}
    \caption{Message passing layers where switches are denoted by red-dashed lines. The node and switch embeddings are represented by blue and red colored blocks respectively, where the number of squares per-block indicates the dimension of the embeddings $h$.}
    \label{fig:mp}
\end{figure}

\subsection{Prediction}
After the $\mathcal{L}$ message passing layers, the embeddings extracted from the input data are used to predict the switch open/close status and a subset of the power flow variables, denoted as independent variables. 

\subsubsection{Variable Space Partition} \label{sec:var_partition}
We partition the variable space into independent and dependent variables. The independent variables constitute the active power flows $p_{ij}$, nodal voltages $v_i$, and switch open/close status $y_{ij}$. The dependent variables constitute the reactive power flows $q_{ij}$, and nodal generation $\{p_i^G, q_i^G\}$. We leverage techniques for variable space reduction to calculate the dependent variables from the independent variables, using constraints \eqref{eq:pbalance}-\eqref{eq:connectivity}. This step ensures that the power physics constraints have certified satisfability, as further discussed in Section~\ref{sec:equality_power_physics}. 

This partition is non-unique. It critically depends on the structure of the given problem which determines the relationship between the sets of variables, and the neural architecture which determines the relationship between inputs, predictions, and consecutive neural layers. \textbf{We further advocate that the neural architecture itself must be physics-informed, to embed domain knowledge and physical constraints directly into the neural network, as we have done in GraPhyR.}

\begin{figure}[!ht]
    \centering
    \includegraphics[width=3.4in,trim={0 0 0 4.5pt}, clip]{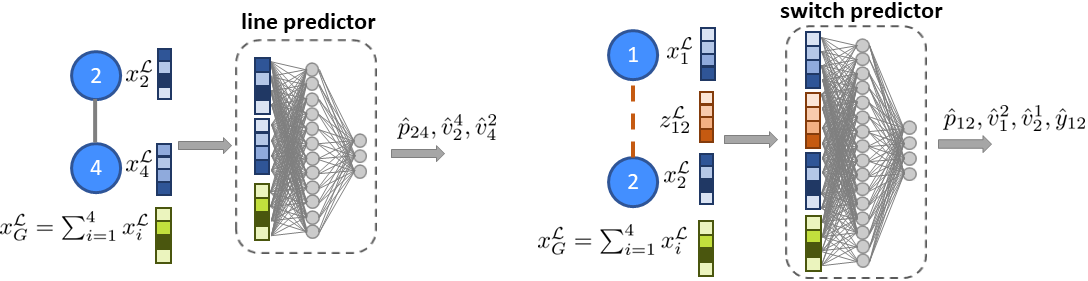}
    \caption{Local predictions made by the switch and line predictors use the node and switch embeddings extracted after $\mathcal{L}$ message passing layers.}
    \label{fig:predictors}
\end{figure}

\subsubsection{Local Prediction Method}
Our prediction method leverages two key observations: (i) the relationship between power flows and voltages are the same for any node-edge pair and are modelled by the physics equations \eqref{eq:pbalance}-\eqref{eq:Ohm_sw}; (ii) the binary nature of switches makes it inherently different from a distribution line. Using these, we define two local prediction methods which use multi-layer perceptrons: a line predictor (L-predictor) and a switch predictor (S-predictor), shown in Fig.~\ref{fig:predictors}. The L-predictor in \eqref{eq:Lpredictor} predicts power flow and the voltages of the two nodes connected by the line using the node and global embeddings. The S-predictor also predicts the probability for the switch to be closed, using the switch embeddings $z_{ij}^\mathcal{L}$ in addition to the node and global embeddings, as in \eqref{eq:Spredictor}. All predictions are denoted with a hat (i.e. $\hat{v}_i$) and will be processed in subsequent layers to render the final topology and dispatch decisions.
\begingroup
\fontsize{9.2pt}{12pt}\selectfont
\begin{align}
    [\hat{p}_{ij}, \hat{v}_i^j, \hat{v}_j^i] &= \text{L-predictor}[x_i^\mathcal{L}, x_j^\mathcal{L}, x_G^\mathcal{L}], \hspace{17pt}\forall (i,j) \in\mathcal{A}\label{eq:Lpredictor}\\
    \hspace{-5pt}[\hat{p}_{ij}, \hat{v}_i^j, \hat{v}_j^i, \hat{y}_{ij}] &= \text{S-predictor}[x_i^\mathcal{L}, x_j^\mathcal{L}, z_{ij}^\mathcal{L}, x_G^\mathcal{L}],\;\forall (i,j) \in\mathcal{A}_{sw} \label{eq:Spredictor}
\end{align}
\endgroup

\normalsize
Our local predictors exploit the full flexibility of GNNs. They are permutation invariant to the input graph data; are independent of the size of the graph (scale-free); and are smaller than the corresponding global predictor for the same grid. The first feature means our framework is robust to changes in input data. The last two features means our framework is lightweight and scalable. This would not be possible with a global predictor which predicts all independent variables from node and switch embeddings across the graph. The size of the input and output layers of a global predictor would depend on the size of the graph and the number of switches, and is the limitation in \cite{haider_SiPhyR}. Table~\ref{tab:predictors_dimension} summarizes the size of local and global predictors for the reconfiguration problem, where $h$ is the dimension of the hidden graph embeddings.

\begin{table}
\centering
\caption{Dimensions of Local and Global Predictors}
\label{tab:predictors_dimension}
\begin{tabular}{lccc}
\toprule[1.5pt]
 & L-predictor & S-predictor & Global-predictor \\
Input dimension & $3h$ & $4h$ & 
$(N+2M_{sw})h$ \\
Output dimension & 3 & 4 & $N+M+2M_{sw}$\\
\bottomrule[1.5pt]
\end{tabular}
\end{table}

\subsection{Voltage Aggregation and Certified Satisfiability of Limits}
The local predictions obtained from the L-predictor and S-predictor generate multiple instances of voltage predictions for each node as indicated by a superscript. Specifically, the number of instances corresponds to the degree of the node $i$, $|\delta_{\mathcal{E}\cup\mathcal{E}_{sw}}(i)|$. We aggregate the voltage predictions to a unique value for each node in the grid as $\tilde{v}_i = \frac{1}{|\delta_{\mathcal{E}\cup\mathcal{E}_{sw}}(i)|}\sum_{j:\{i,j\}\in\mathcal{E}\cup\mathcal{E}_{sw}} \hat{v}_i^j$. The voltage predictions are then scaled onto the box constraints \eqref{eq:v_pcc} with $v_i = \underline{v} \cdot (1 - \tilde{v}_i) + \overline{v} \cdot \tilde{v}_i$. Notably, by selecting voltages as an independent variable in our variable space partition, we certify that voltage limits across the grid will always be satisfied, a critical aspect of power systems operation.

\subsection{Topology Selection using Physics-Informed Rounding}\label{sec:topo_selec_phyr}
The S-predictor provides probabilistic predictions for open/close decisions of each switch. We recover binary decisions using a physics-informed rounding (PhyR) algorithm \cite{haider_SiPhyR}. We exploit the radiality of distribution grids, which requires $\mathcal{S}=N-1-M$ switches to be closed so there are always $N-1$ conducting lines. The PhyR method selects the $\mathcal{S}$ switches with the largest probabilities $\hat{y}_{ij}$ and closes them by setting the corresponding $y_{ij}=1$; the remaining switches are opened, $y_{ij}=0$. This enforces \eqref{eq:Ohm_sw} and \eqref{eq:radial}. Note that as distribution grid technologies advance, bidirectional and loop flows may be easily incorporated in new protection schemes. This would remove the radiality constraint, which GraPhyR can accommodate with suitable modifications to PhyR.

A note must be made about the practical implementation: PhyR is implemented with $\min$ and $\max$ operators which return gradients of $0$, ``killing'' the gradient information necessary for backpropagation. We preserve these gradients in the computational graph by setting all but one switches to binary values, those with $\mathcal{S}-1$ largest probabilities. Training guides the remaining switch towards a binary value.

\subsection{Certified Satisfiability of Power Physics} \label{sec:equality_power_physics}
The final neural layer recovers the full variable space and enforces power flow constraints through open switches. 
The following steps happen sequentially: 
\begin{enumerate}
    \item Given $\mathbf{y}$ and the independent variables we compute the reactive power flows $\tilde{q}_{ij}$ using \eqref{eq:Ohm}-\eqref{eq:Ohm_sw}.
    \item Given $\mathbf{y}$ we enforce \eqref{eq:sw_p} and \eqref{eq:sw_q} as $p_{ij} = (\hat{p}_{ij} - 0.5) \cdot 2\mathcal{M}y_{ij}$ and $q_{ij} = (\tilde{q}_{ij} - 0.5) \cdot 2\mathcal{M}y_{ij}$ respectively. By explicitly setting flows through open switches to zero we enforce the constraints in a hard way. This step also explicitly enforces the condition constraint describing Ohm's Law \eqref{eq:Ohm_sw}.
    \item Active and reactive power nodal generation is calculated using \eqref{eq:pbalance} and \eqref{eq:qbalance}, respectively.
\end{enumerate}

\subsection{Loss Function}
The neural network learns to optimize by using an unsupervised framework. It has two objectives: to minimize line losses in \eqref{eq:objectivefnc}, and to minimize inequality constraint violations of generation constraint \eqref{eq:var_lims} and connectivity constraints \eqref{eq:connectivity}. Denoting these constraints as $h(\mathbf{x}, \boldsymbol{\psi})\leq 0$, we regularize the loss function using a soft-loss penalty with hyperparameter $\lambda$. The loss function is $l = f(\mathbf{x}, \boldsymbol{\psi}) + \lambda ||\text{max}\{0, h(\mathbf{x}, \boldsymbol{\psi})\}||_2$.

\textit{Remark 1:} The inequality constraints \eqref{eq:sw_p}, \eqref{eq:sw_q}, and \eqref{eq:v_pcc} have certified satisfiability by design of GraPhyR.

\textit{Remark 2:} Our loss function is unsupervised, and does not need the optimal solutions, which may be unknown or computationally prohibitive to compute. 

\section{Experimental Results} \label{sec:results}
\subsection{Dataset and Experiment Setup}
We first evaluate GraPhyR on a canonical distribution grid BW-33 \cite{network_rec} with 33 nodes, 29 lines, and 8 switches. We generate a variant of BW-33, called $\mathcal{G}_1$ with 33 nodes, 27 lines, and 10 switches. Second, we evaluate GraPhyR on a a model of a real distribution grid TPC-94 \cite{Su_2005_83dataset} with 94 nodes and 14 switches. The grid has 11 individual distribution feeders which can be connected to one another via switches (i.e. reconfiguration) to share load across feeders. We use the datasets from \cite{haider_SiPhyR} which introduce distributed solar generation in the grid with a penetration of $25\%$ generation-to-peak load. Loads are perturbed about their nominal value as typically done in literature. The two networks are shown in Fig.~\ref{fig:grids}. The dataset has 8600 data points per grid which are divided as $80/10/10$ for training/validation/testing.

\begin{figure}[!t]
\centering
\subfloat[]{\includegraphics[scale=0.2,trim={0 1.2cm 0 1.3cm},clip]{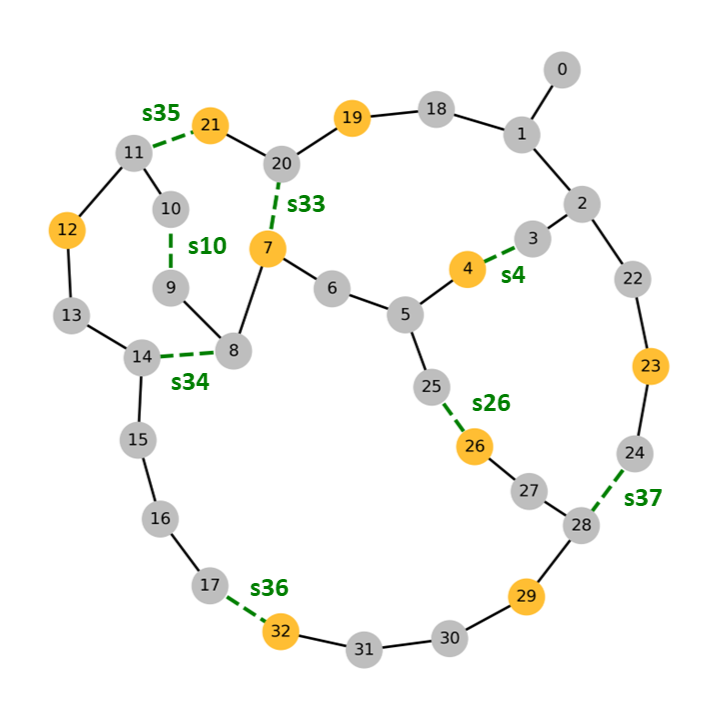}}
\hfill
\subfloat[]{\includegraphics[scale=0.2,trim={0 1.4cm 0 1.2cm},clip]{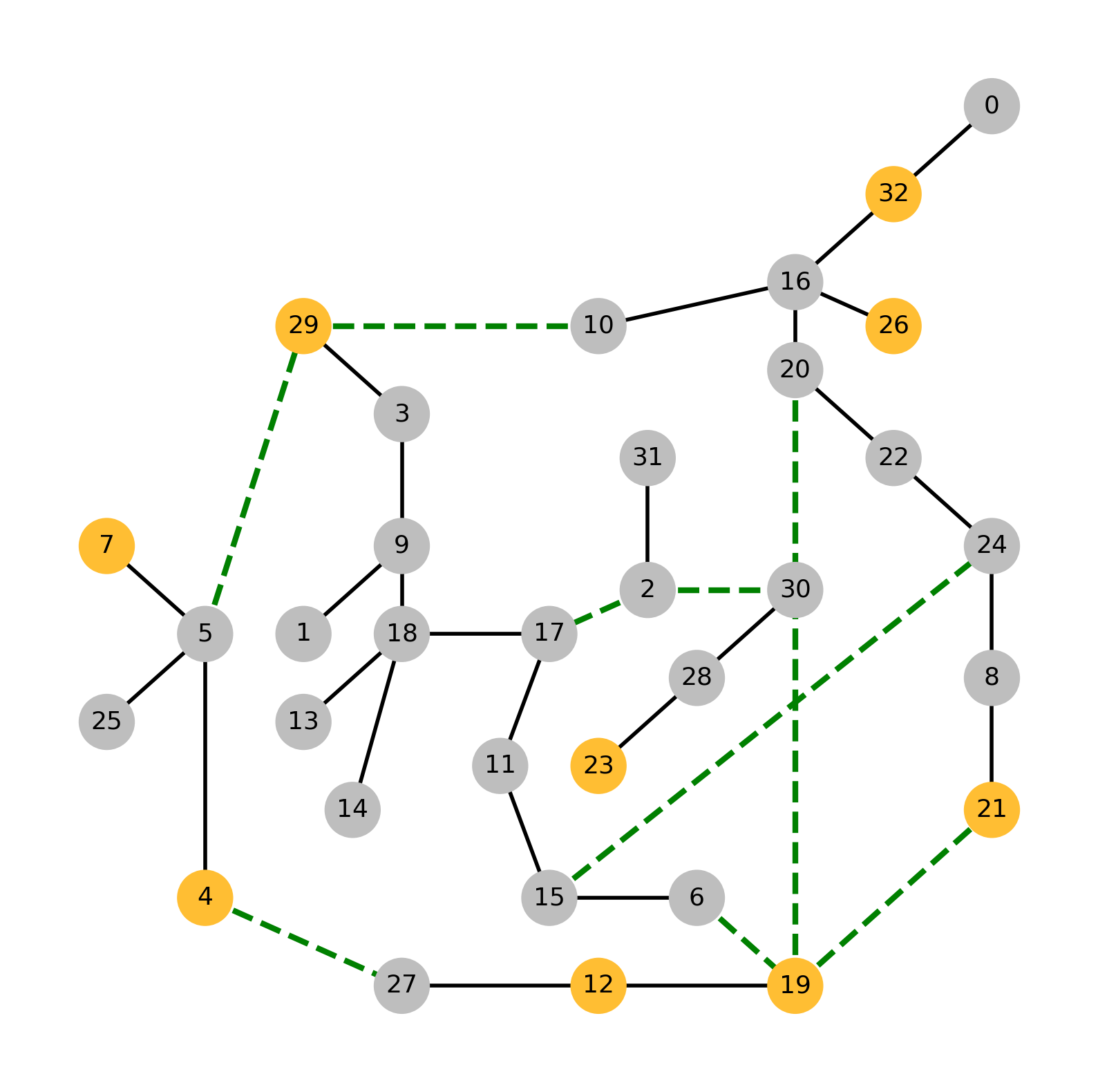}}%
\caption{Grid topology of BW-33 (left) and the synthetic $\mathcal{G}_1$ (right). Switches indicated with green dashed lines. Solar generator locations indicated with yellow nodes.}
\label{fig:grids}
\end{figure}

We implement GraPhyR using PyTorch and train on the MIT supercloud \cite{supercloud}. GraPhyR has $4$ message passing layers each with dimension $8$ ($\mathcal{L}=4, h=8$). The L-predictor and S-predictor have a single hidden layer with dimension 24 and 32 respectively. We use $10\%$ dropout, batch normalization, and ReLU activation in both predictors. The soft loss hyperparameter is $\lambda=100$, big-$\mathcal{M}$ relaxation parameter is $0.5$ per unit (p.u.). The voltage bounds for BW-33 are $\underline{v}=0.83,\overline{v}=1.05$ p.u. to adapt to the lossy behaviour of the grid \cite{network_rec,haider_SiPhyR}, and for TPC-94 we use the typical $\underline{v}=0.83,\overline{v}=1.05$ p.u. limits.
We use ADAM optimizer with a learning rate of $\gamma = 5e^{-4}$, a batch size of $200$, and train for 1500 epochs. We evaluate the performance of the neural framework using a committee of networks approach. We train 10 models with independent weight initialization and average the predictions across all models. 

\subsection{Performance Metrics}
We adopt the performance metrics defined in \cite{haider_SiPhyR} to assess prediction performance. The asterisks notation (i.e. $v^\ast$) denotes the optimal solution obtain from a MIP solver.

\noindent \textbf{Dispatch error:} optimality metric of mean-squared error (MSE) in optimal generator dispatch: $\frac{1}{N}\sum_{j\in \mathcal{N}}{(p_j^G - p_j^{G\ast})^2 + (q_j^G - q_j^{G\ast})^2}$.

\noindent \textbf{Voltage error (VoltErr):} optimality metric of MSE in nodal voltage prediction: $\frac{1}{N}\sum_{j\in \mathcal{N}}{(v_j - v_j^{\ast})^2}$.

\noindent \textbf{Topology error:} optimality metric of the Hamming distance \cite{Hamming_1986} between two topologies, calculated as the ratio of switch decisions not in the optimal position: $\frac{1}{M_{sw}}\sum_{(i,j)\in\mathcal{A}_{sw}}{(y_{ij}-y_{ij}^\ast)^2}$.

\noindent \textbf{Inequality violation:} feasibility metric of the magnitude of violations in constraint set, measuring the mean and maximum as $\frac{1}{|h|}\sum_{k}{\max{\{ 0, h^k(\mathbf{x}, \boldsymbol{\psi}) \}}}$ and $\max_k{\{\max{\{ 0, h^k(\mathbf{x}, \boldsymbol{\psi}) \}}\}}$.

\noindent \textbf{Number of violations exceeding a threshold:} feasibility metric of the number of inequality constraints which are violated by more than an $\epsilon$ threshold: $\sum_{k}{\mathbb{I}_{\max{\{ 0, h^k(\mathbf{x}, \boldsymbol{\psi}) \}} > \epsilon}}$.

\begin{table*}[t]
\centering
\caption{Simulation Results on the BW-33, $\mathcal{G}_1$, and TPC-94 grids tested on 860/8640 instances. Lower values are better for all Metrics.}
\label{tab:results}
\begin{tabular}{c l c c c c c c c} 
\toprule[1.5pt]
  & & \multicolumn{6}{c}{\textbf{Metric}}\\
  & \textbf{Method} & Dispatch error (MSE) & Voltage error (MSE) & Topology error & Ineq Viol (mean) & Ineq viol (max) & Num ineq viol $>0.01$\\ 
  \cline{0-1}\\
  & GraPhyR (our method) & 2.22e-03 & 5.55e-03 & 39.6\% & 2.33e-03 & 2.74e-02 & 20.8 \\
  & Global-GraPhyR & 1.93e-04 & 1.03e-03 & 38.8\% & 9.86e-04 & 2.00e-02 & 4.1\\
  (a) & SiPhyR \cite{haider_SiPhyR} & 2.89e-02 & 1.69e-03 & 41.5\% & 4.79e-04 & 4.23e-02 & 5.72\\
  & InSi \cite{haider_SiPhyR} & 3.24\text{e-}02 & 2.30\text{e-}03 & 49.7\% & 1.53\text{e-}03 & 0.148 & 16.3 \\
  & Semi-supervised GraPhyR & 2.48e-03 & 5.66e-03 & 1.33\% & 2.45e-03 & 3.48e-02 & 21.2 \\
  & Supervised-SiPhyR \cite{haider_SiPhyR} & 5.78\text{e-}04 & 1.35\text{e-}03 & 33.6\% & 5.49\text{e-}04 & 4.84\text{e-}02 & 7.26 \\
  \hline
  (b) & GraPhyR [BW-33, $\mathcal{G}_1$] & 2.64e-03 & 8.37e-03 & 42.4\% & 2.47e-03 & 2.95e-02 & 21.9 \\
  \hline
  &sw10 closed GraPhyR & 2.48e-03 & 8.36e-03 & 31.7\% & 3.46e-03 & 7.89e-02 & 28.6 \\
   & sw35 closed GraPhyR & 2.53e-03 & 5.57e-03 & 37.4\% & 3.78e-03 & 7.57e-02 & 32.5 \\
  (c) &sw36 closed GraPhyR & 2.37e-03 & 1.61e-02 & 45.0\% & 3.39e-03 & 5.98e-02 & 27.9 \\
  &sw10 opened GraPhyR & 3.90e-03 & 5.44e-03 & 32.5\% & 7.06e-03 & 1.75e-01 & 37.2 \\
  & sw35 opened GraPhyR & 3.45e-03 & 8.46e-03 & 53.9\% & 6.44e-03 & 1.50e-01 & 37.3 \\
  &sw36 opened GraPhyR & 4.25e-03 & 5.64e-03 & 28.2\% & 7.35e-03 & 1.71e-01 & 38.8 \\
  \hline
   & GraPhyR [TCP-94] & 1.57e-02 & 1.49e-02 & 41.9\% & 1.13e-02 & 2.56e-01 & 140\\
   (d) & SiPhyR [TCP-94] \cite{haider_SiPhyR} & 1.12e-02 & 1.51e-02 & 45.4\% & 7.27e-04 & 4.25e-02 & 37.5\\
   & InSi [TCP-94] \cite{haider_SiPhyR} & 1.41 & 3.31e-02 & 44.3\% & 3.21e-02 & 3.01 & 162\\
  \bottomrule[1.5pt]
\end{tabular}
\end{table*}

\subsection{Benchmark models}
We benchmark our results against multiple solution techniques, including a traditional MIP solver and different ML frameworks, as described below.
\newline 
\noindent \textbf{Optimizer}: Traditional optimization solver, Gurobi, a state-of-art commercial solver for MIPs. \newline 
\noindent \textbf{GraPhyR}: The proposed method described in Section~\ref{sec:graphyr}, with a GNN with switch embeddings, local predictors (the L-predictor and S-predictor), and PhyR layer to recover binary decisions. 
\newline 
\noindent \textbf{Global-GraPhyR}: A modification of the proposed GraPhyR method which uses a global predictor which predicts all independent variables from node and switch embeddings.
\newline 
\noindent \textbf{SiPhyR}: An physics-informed method introduced in \cite{haider_SiPhyR} which uses a lightweight fully connected neural network with a sigmoidal output layer to predict the independent variables (rather than the GNN). The topology selection uses the PhyR algorithm, and the variable space decomposition is modified to additionally include integer variables indicating the directionality of power flow in the lines.
\newline 
\noindent \textbf{InSi}: A simple neural network without the proposed PhyR layer \cite{haider_SiPhyR}. Integer solutions for the switch status are encouraged (read: not enforced) by using a differentiable relaxation of the step function, the integer sigmoid (InSi): $\sigma_{InSi}(z) = \left[ 2\frac{1+\mu}{\mu+e^{-\tau z}} -1 \right]_{+}$, where $\tau, \mu$ are free parameters \cite{cao_sigmoidal}.

\subsection{Case (a). GraPhyR with Local vs. Global Predictors}
We first compare GraPhyR with local predictors to a variant with a global predictor, termed Global-GraPhyR. The global predictor determines all independent variables (real power flows, voltages, switch probabilities) using all node and line embeddings. We implement the global predictor with a single hidden layer of the same size as the input dimension. The global predictor has input/output dimensions of 328/78 as compared to the L-predictor and S-predictor with dimensions of 24/3 and 32/4 respectively. Note that the global predictor predicts one voltage per node so voltage aggregation is not needed. We also compare the performance of GraPhyR with that of prior work which use a simple neural network with two hidden layers \cite{haider_SiPhyR}: \textit{SiPhyR} which employs PhyR; and \textit{InSi} which approximates a step function.


Table~\ref{tab:results}-(a) shows the prediction performance for these methods. We first observe that \textbf{the GNN frameworks achieve lower dispatch error}, with Global-GraPhyR outperforming SiPhyR by two orders of magnitude. The GNN uses topological information to optimize the dispatch and satisfy loads. Second, \textbf{the PhyR-based frameworks achieve lower topology errors} by up to 10\%, by embedding the discrete decisions directly within the ML framework. However, the topology error remains high ($>30\%$), demonstrating the challenge in learning to optimize this combinatorial task. Finally, \textbf{SiPhyR and Global-GraPhyR achieve the best performance across feasibility metrics}, with lower magnitude and number of inequality violations. Notably, the maximum inequality violation is an order of magnitude higher for InSi which does not benefit from PhyR, and GraPhyR which makes local predictions. This is expected. First, PhyR explicitly accounts for binary variables within the training loop to enable the end-to-end learning: PhyR selects a feasible topology upon which the neural framework predicts a near-feasible power flow solution. Second, GraPhyR sacrifices some prediction performance for the flexibility to train and predict on multiple graphs. Figure~\ref{fig:ineq_viol} plots the mean inequality violation of the 10 trained GraPhyR models, for the sets of inequality constraints, namely \eqref{eq:var_lims}, \eqref{eq:v_pcc}, and \eqref{eq:connectivity}. The constraints are always respected for voltage (by design) and connectivity (by constraint penalty). Nodal generation constraints are frequently violated as the lowest cost (lowest line losses) solution is to supply all loads locally.

We next test the limits of topology prediction within our ML framework by comparing with a semi-supervised approach. The loss function includes a penalty on the switch status:

\begin{align}
    l_{sm} &= f(\mathbf{x}, \boldsymbol{\psi}) + \lambda ||\text{max}\{0, h(\mathbf{x}, \boldsymbol{\psi})\}||_2 + \textcolor{blue}{\mu \|\mathbf{y} - \mathbf{y}^* \|_2} \label{eq:semi_loss_fnc}
\end{align}

Table~\ref{tab:results}-(a) shows the performance of the semi-supervised GraPhyR. We also include results of Supervised-SiPhyR from \cite{haider_SiPhyR} which uses a regression loss for voltages, generation, and switch statuses, and an inequality constraint violation penalty:

\begin{align}
    l_{sup}(z,\varphi) &= \textcolor{blue}{\Vert {(\mathbf{v} - \mathbf{v^{\ast}})^2 + (\mathbf{p^G} - \mathbf{p^{G\ast}})^2 + (\mathbf{q^G} - \mathbf{q^{G\ast}})^2}\Vert_2^2} \nonumber \\
     & \textcolor{blue}{+ \Vert {(\mathbf{y}-\mathbf{y^\ast})^2}\Vert_2^2} + \lambda ||\text{max}\{0, h(\mathbf{x}, \boldsymbol{\psi})\}||_2
\end{align}

The results show that Semi-supervised GraPhyR outperforms Supervised-SiPhyR on topology error, achieving near-zero error. This substantial difference can be attributed to the GNN which embeds topological data directly within the framework. Although these (semi-)supervised approaches achieve good performance, they are not practicable. They require access to the optimal solutions, which may be computationally prohibitive to generate across thousands of training data points.

A note must be made on computational time. Solving the DyR problem using Gurobi takes on average 201 milliseconds for BW-33, and 18 seconds for a 205-node grid per instance. Actual computational times vary significantly with varying load conditions which stress grid voltages (Ex. 17-fold increase for the 205-node grid during high load periods \cite{haider_SiPhyR}). In contrast, the inference time of GraPhyR is only 84 milliseconds for a batch of 200 instances.

\begin{figure}[!ht]
    \centering
    \includegraphics[width=2.3in, trim={0 0.25cm 0 0.25cm}, clip]{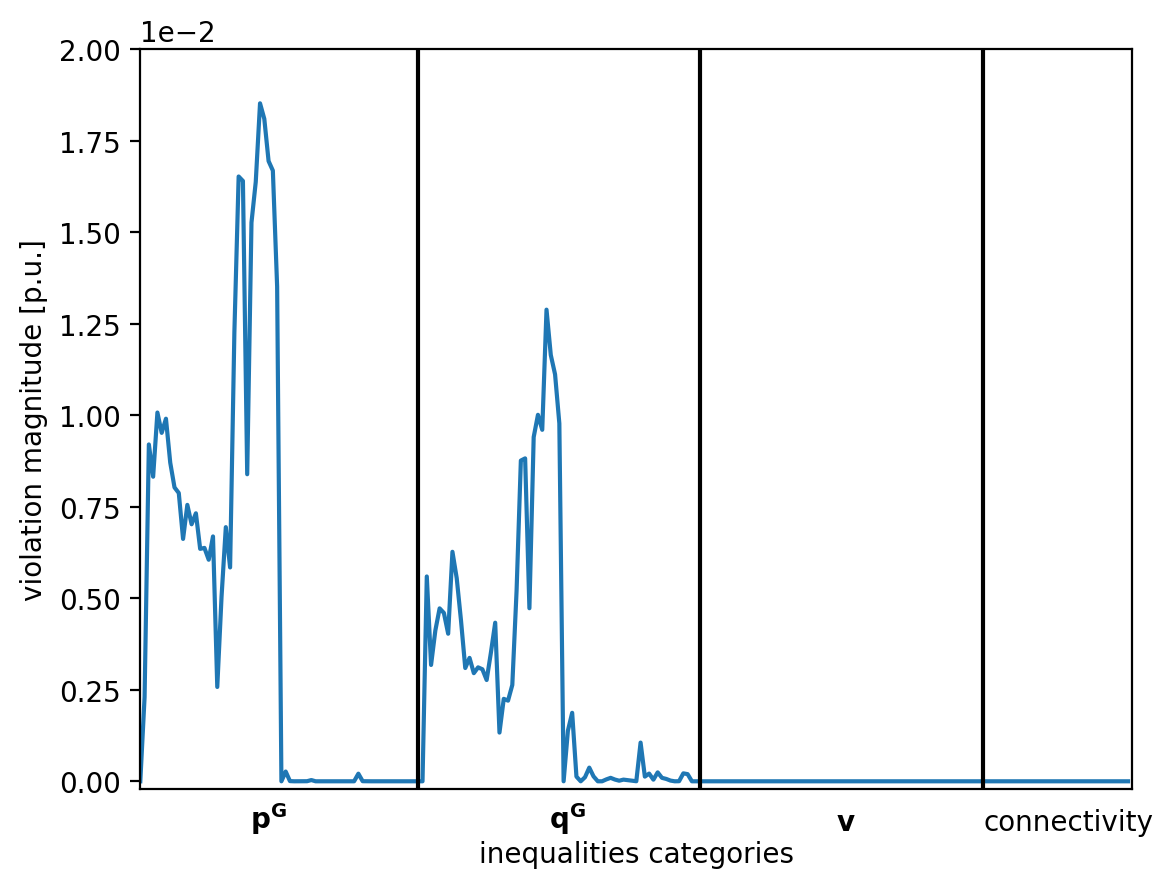}
    \caption{Magnitude of the inequality constraint violations for GraPhyR. The constraint sets on nodal generation ($\mathbf{p^G} \text{ and } \mathbf{q^G}$, as in Eq.~\eqref{eq:var_lims}), voltage limits ($\mathbf{v}$, as in Eq.~\eqref{eq:v_pcc}), and connectivity constraints (Eq.~\eqref{eq:connectivity}) are separated by black vertical lines.}
    \label{fig:ineq_viol}
\end{figure}

\subsection{Case (b). Prediction Performance on Multiple Grids}
A key feature of GraPhyR is its ability to solve the DyR problem across multiple grid topologies. We construct a dataset using problem instances from both the BW-33 and $\mathcal{G}_1$ grids in Fig.~\ref{fig:grids}. These grids have the same number of nodes, but different topologies, number of lines and switches, and location of switches. We train and test GraPhyR on this dataset. Table~\ref{tab:results}-(b) shows these results, showing the average errors and inequality constraint violations across problem instances of both grids. The performance of GraPhyR on the two grids is similar to that of GraPhyR on a single grid, showing that GraPhyR can learn the power flow representation across multiple topologies and across multiple grids at the same time.

\subsection{Case (c). Adapting to Changing Grid Conditions}
We next test GraPhyR on changing grid conditions, such as (un)planned maintenance by the grid operator or switch failure. Since power flows are highly correlated with the grid topology, changes in the set of feasible topologies due to maintenance or equipment failure can significantly change the prediction accuracy. Rather than training on multiple scenarios, we train only on the BW-33 grid for normal operating conditions and test on cases where a switch is required to be open or closed. 

Results are shown in Table~\ref{tab:results}-(c). Generally the dispatch error, voltage error, and average inequality violation magnitudes remain similar to cases of normal operation. However, there is a notable increase in the number of inequality violations, and when forcing a switch open, an order of magnitude increase in the maximum inequality violations. Forcing a switch open removes an edge from the GNN graph. The resulting graph is more sparse, reducing access to information during message passing and changing the information contained in the node and switch embeddings.

The topology error is more nuanced. When switch 36 is closed, there is an increase in voltage and topology error. This is because without any operator requirements on switch statuses, switch 36 remains optimally open for all load conditions. Thus, when switch 36 is required to be open, there is a significant decrease in topology error, by almost 10\%. Since we did not trained on other scenarios, GraPhyR struggles to optimize the topology and predict voltages when the grid conditions deviate significantly from the training data -- such as when switch 36 is closed. Similar performance degradation happens when switch 35 is required to be open; this switch is optimally closed for all load conditions. Interestingly, the status of switch 10 (open or closed) does not affect the topology error, although this switch is typically closed in the training data. There may be multiple (near-)optimal topologies with similar objective value. Regularizing the dataset or performance metrics against these multiple solutions may be necessary to improve prediction performance. 

\subsection{Case (d). GraPhyR on a larger grid}
We test GraPhyR on the larger TPC-94 grid, using the same GNN parameters ($\mathcal{L}=4, h=8$) as in BW-33. Compared to the SiPhyR framework, our GraPhyR approach has comparable performance on optimality metrics and slightly lower topology error. The inequality constraint violations are higher than SiPhyR, but lower than InSi. This result is expected, since the GNN parameters were kept the same as the for $BW-33$ while the complexity of the problem increased drastically with the grid size and the number of switches. In comparison, the results for SiPhyR and InSi use a larger neural network for TPC-94 than for BW-33. Parameter tuning of the GNN should be done for each network.

\subsection{Utility perspective on DyR}
A typical distribution substation may have two to nine distribution feeders per substation. A feeder model may consist of 60 nodes for a medium-sized feeder. The resulting distribution grid model for a single substation may consist of 200-600 nodes. Larger grids may exist in urban load centers. A typical distribution grid may have 10-40 switches (two to nine feeders, with four switches per feeder). These consist of both normally closed switches (NCS) which connect feeders to the substation and across feeders, and normally open switches (NOS) which are typically used for fault location, isolation, and service restoration (FLISR) activities. In this work, we assume both NCS and NOS are available for optimization through DyR. Our GraPhyR framework enables an automated approach to DyR, that can reduce losses as compared to a rule-based approach by 2.5\% \cite{haider_SiPhyR}; for a utility with a 100MW peak load, this translates to savings of US\$200,000 per year \cite{CVR-1}.

\section{Conclusion} \label{sec:conclusion}
We developed GraPhyR, an end-to-end physics-informed Graph Neural Network framework to solve the dynamic reconfiguration problem. We model switches as gates in the GNN message passing, embed discrete decisions directly within the framework, and use local predictors to provide scalable predictions. Our simulation results show GraPhyR outperforms methods without GNNs in learning to predict optimal solutions, and offers significant speed-up compared to traditional MIP solvers. Further, our approach adapts to unseen grid conditions, enabling real-world deployment. Future work will investigate the scalability of GraPhyR to larger grids (200+ nodes), approaches to reduce inequality constraint violations, and regularization strategies to improve topology prediction. Finally, further efforts are needed in developing good datasets with representative timeseries data in distribution grids. 

\bibliographystyle{IEEEtran}
\bibliography{bibliography}

\end{document}